\documentclass{article}

\usepackage[numbers]{natbib}

\usepackage[final]{neurips_2024}

\usepackage[utf8]{inputenc} 
\usepackage[T1]{fontenc}    
\usepackage{hyperref}       
\usepackage{url}            
\usepackage{booktabs}       
\usepackage{amsfonts}       
\usepackage{nicefrac}       
\usepackage{microtype}      
\usepackage{xcolor}         

\usepackage{multirow}
\usepackage{wrapfig} 
\usepackage{makecell}
\usepackage{subfig}
\usepackage{caption}
\usepackage{blindtext}
\usepackage{comment}
\usepackage{graphicx}
\usepackage{lipsum}
\usepackage{colortbl}

\newcommand{\ie}{\textit{i}.\textit{e}., }
\newcommand{\eg}{\textit{e}.\textit{g}., }

\usepackage{bm}
\usepackage{xcolor}
\usepackage{textcomp}
\usepackage{amssymb} 
\usepackage{amsmath}
\usepackage{wrapfig}
\usepackage{newfloat}
\usepackage{tablefootnote}
\usepackage{booktabs}
\usepackage{mathtools}
\usepackage{inconsolata}
\usepackage{enumitem}

\usepackage{algpseudocode}
\usepackage{algorithm}

\title{TPC: Test-time Procrustes Calibration for Diffusion-based Human Image Animation}

\author{
  Sunjae Yoon \quad  Gwanhyeong Koo\quad Younghwan Lee \quad Chang D. Yoo\thanks{Corresponding author} \\
  Korea Advanced Institute of Science and Technology (KAIST) \\
  \texttt{\{sunjae.yoon,cd\_yoo\}@kaist.ac.kr} \\
}

\begin{document}

\maketitle

\begin{abstract}
  Human image animation aims to generate a human motion video from the inputs of a reference human image and a target motion video. Current diffusion-based image animation systems exhibit high precision in transferring human identity into targeted motion, yet they still exhibit irregular quality in their outputs. Their optimal precision is achieved only when the physical compositions (i.e., scale and rotation) of the human shapes in the reference image and target pose frame are aligned. In the absence of such alignment, there is a noticeable decline in fidelity and consistency. Especially, in real-world environments, this compositional misalignment commonly occurs, posing significant challenges to the practical usage of current systems. To this end, we propose Test-time Procrustes Calibration (TPC), which enhances the robustness of diffusion-based image animation systems by maintaining optimal performance even when faced with compositional misalignment, effectively addressing real-world scenarios. The TPC provides a calibrated reference image for the diffusion model, enhancing its capability to understand the correspondence between human shapes in the reference and target images. Our method is simple and can be applied to any diffusion-based image animation system in a model-agnostic manner, improving the effectiveness at test time without additional training. The project is available at \href{https://github.com/dbstjswo505/TPC}{\texttt{github.com/dbstjswo505/TPC}}
\end{abstract}
\section{Introduction}
%
Denoising diffusion models \cite{dhariwal2021diffusion,song2020score,song2020denoising,ho2020denoising} have transformed the generative landscape of artificial intelligence, leading to groundbreaking achievements \cite{koo2024flexiedit,popov2021grad,yoon2024frag,yoon2024dni} in image, speech, and video generation.
We explore the application of diffusion model in the specific context of image-to-video generation, focusing on the task of human image animation.
The technology of image animation holds great promise, enabling immersive and interactive experiences in entertainment, virtual reality, and digital communication.
The human image animation systems \cite{xu2023magicanimate,wang2023disco,karras2023dreampose,hu2023animate} are designed to work with referential human image and target motion video, where they transfer the human identity into the target motion, ensuring seamless and unobtrusive integration.
This process requires understanding the correspondence between human shapes in the reference image and the target pose frame.
%
%

\begin{figure*}[t!]
\centering
   \includegraphics[width=1.0\textwidth]{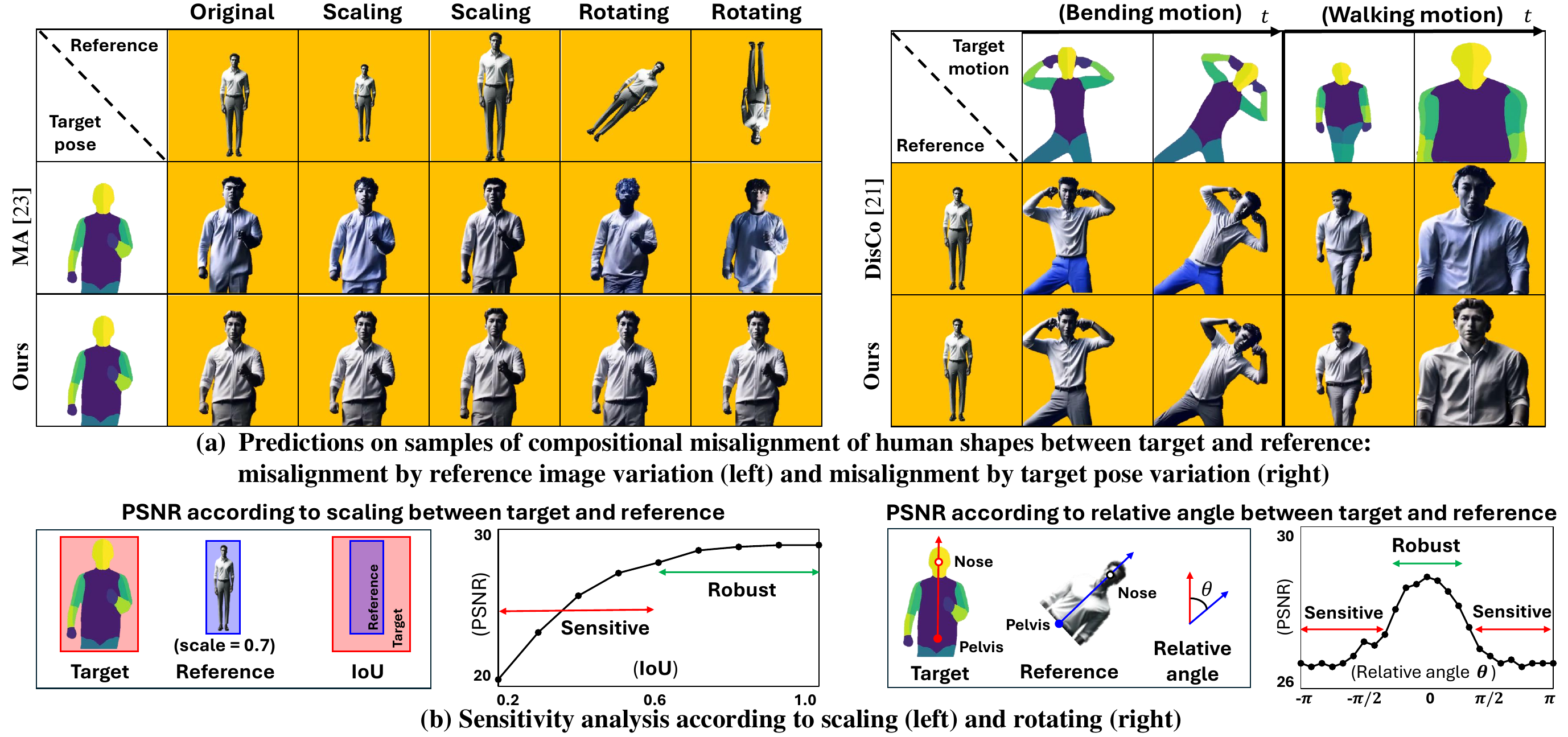}
   \caption{Illustration of compositional misalignment: (a) Results of current human image animation models \cite{xu2023magicanimate,wang2023disco} on samples in compositional misalignment of human shapes between reference and target. (b) Sensitivity analysis according to variation of compositional misalignment by scaling and rotating (MA: MagicAnimate). Best viewed with zoom.}
\label{fig:observation_evidence}
\vskip -0.2in
\end{figure*}

%
Recent advancements \cite{xu2023magicanimate,wang2023disco} of human image animation systems have demonstrated notable precision in adapting human identity into target motion.
%
%
Despite advancements, these systems continue to suffer from irregular quality of image animation when the compositions (\ie scale and rotation) of human shapes are not aligned between reference image and target motion.
%
%
%
To be specific, Figure \ref{fig:observation_evidence} (a) presents exploratory experiments on the compositional misalignment of human shapes between the reference and target.
As shown in the left experiments, for a given target pose, adjusting the composition of the same human in the reference image (\ie via scaling or rotating) results in inconsistent and low-fidelity output images, especially in terms of clothes and faces of the human.
%
%
%
%
%
%
%
Furthermore, in the right experiments, when providing motion sequences that display various dynamic movements for a given reference image, the human image animation outputs consistently show low fidelity, especially evident in target poses (\eg bending or approaching close to viewpoint) that cause significant differences in the composition of the human shape.
To quantitatively investigate these, Figure \ref{fig:observation_evidence} (b) presents a sensitivity analysis evaluating how current image animation systems \cite{xu2023magicanimate, wang2023disco} respond to varying degrees of the compositional misalignment of human shape.
The left shows the fidelity (\ie PSNR) of the resulting human according to the relative scale difference of input human shapes between the target and reference.
We employ the Intersection of Union (IoU) of bounding boxes of the shapes for relative scale.
The right shows fidelity according to the variations of relative angle $\theta$ \footnote{We define the axis of the torso of a human from the pelvis to nose using estimated human body key points \cite{cao2017realtime} and measure the relative angle of axes between target and reference.} of human shapes between the target and reference.
%
%
Current systems demonstrate significant vulnerability to the variations of the compositions, indicating that output fidelity forms a robust region only in the areas with compositionally aligned conditions (\ie $-\frac{\pi}{6}<\theta<\frac{\pi}{6}$, IoU $>0.7$).

\begin{wrapfigure}{r}{0.5\textwidth}
    \vskip -0.2in
    \centering    
    \includegraphics[width=0.5\textwidth]{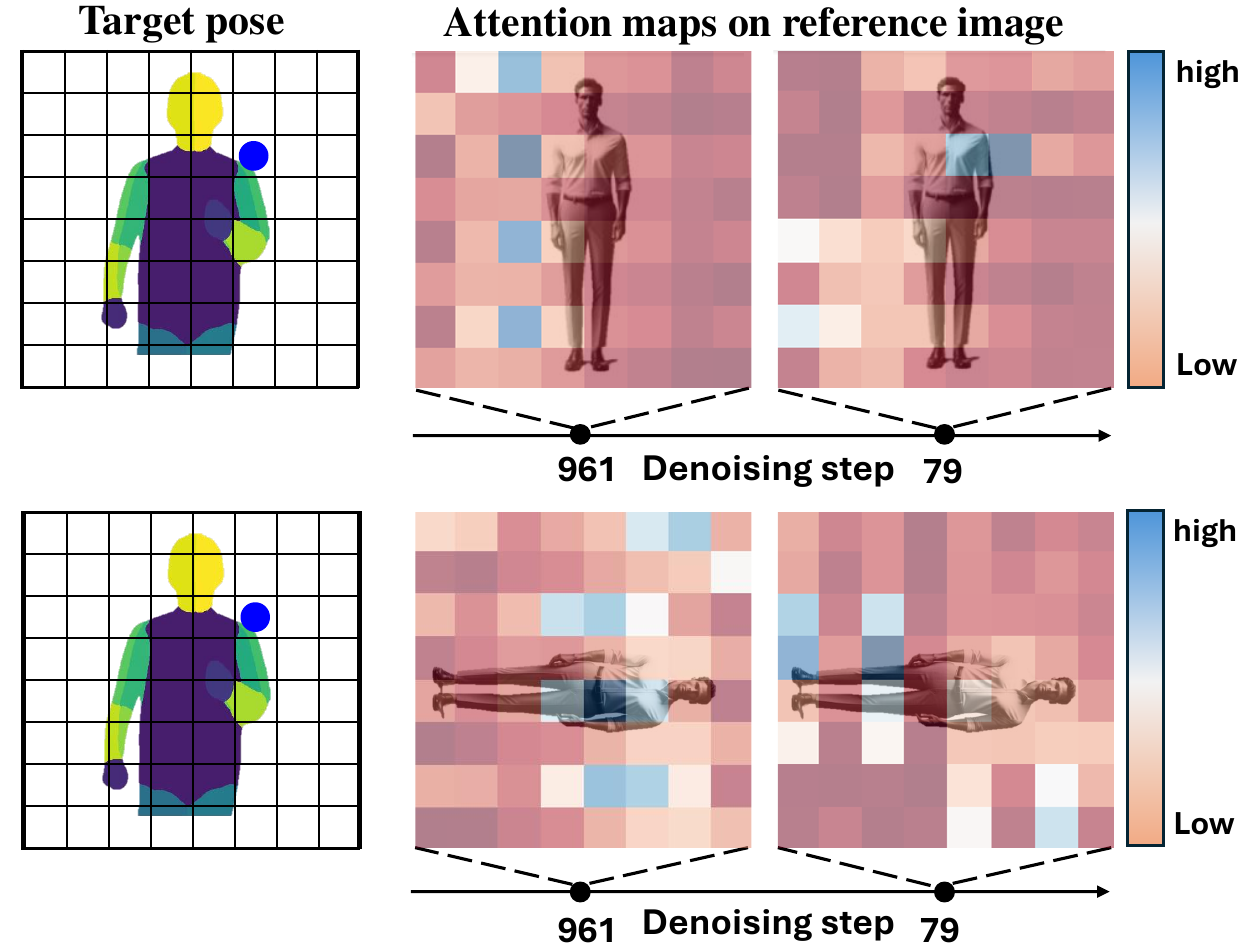}
    \caption{Attention maps on the reference image corresponding to the target human shape (\eg shoulder at blue point) according to denoising.}
    \label{fig:attmap}
    \vskip -0.1in
\end{wrapfigure}
%
%
%
%
In fact, diffusion-based systems are inevitably susceptible to this compositional misalignment.
%
%
%
%
The diffusion model uses a reference human image as a condition to generate controlled output from noise based on the target pose. 
Here, the conditioning is performed through cross-attention based on a visual similarity between patch-wise features of the target pose frame and reference image.
%
%
To be specific, Figure \ref{fig:attmap} shows the attention map about a single patch of the target frame (\ie shoulder denoted by blue point) from the reference image during the denoising process.
%
%
The upper section displays a sample where the human shapes are relatively aligned between the target and the reference, while the lower section shows a case where this alignment is not present.
%
%
Initially, the attention maps were blurry in both cases. However, as denoising continued, it became clear that samples with aligned human shapes correctly established a correspondence between the target frame and the reference image. (\ie Attention to the shoulder on the target frame is focused on the shoulder on the reference image.)
%
%
However, when the shape is misaligned, attention to the target frame's shoulder is incorrectly focused on unrelated areas in reference, even up to the last stages of denoising.
%
%
Consequently, the compositional misalignment between the target and the reference image hinders accurate attention correspondence throughout the denoising process.
\begin{figure*}[t!]
\centering
   \includegraphics[width=1.0\textwidth]{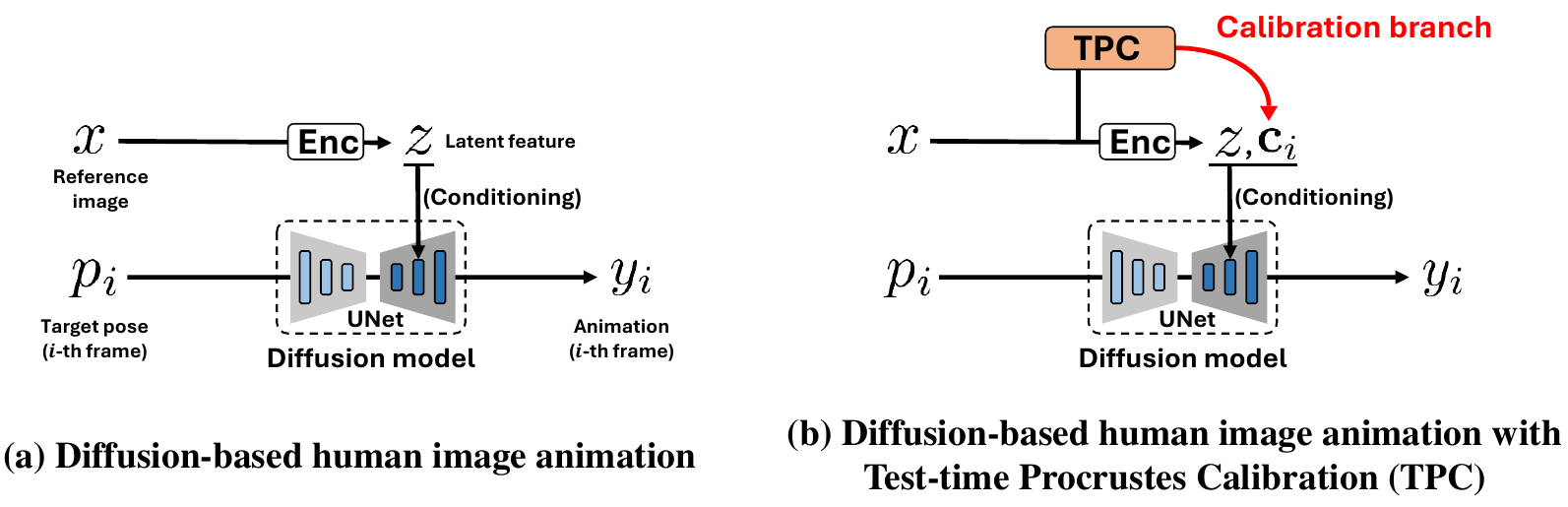}
   \caption{Illustration of (a) current diffusion-based human image animation systems and (b) Test-time Procrustes Calibration (TPC) on top of these systems. The TPC can be applied to diffusion-based models in a model-agnostic manner, enhancing the fidelity and consistency of the output video.}
\label{fig:pc}
\vskip -0.2in
\end{figure*}

\begin{wrapfigure}{r}{0.5\textwidth}
    \centering
    \vskip -0.15in
    \includegraphics[width=0.5\textwidth]{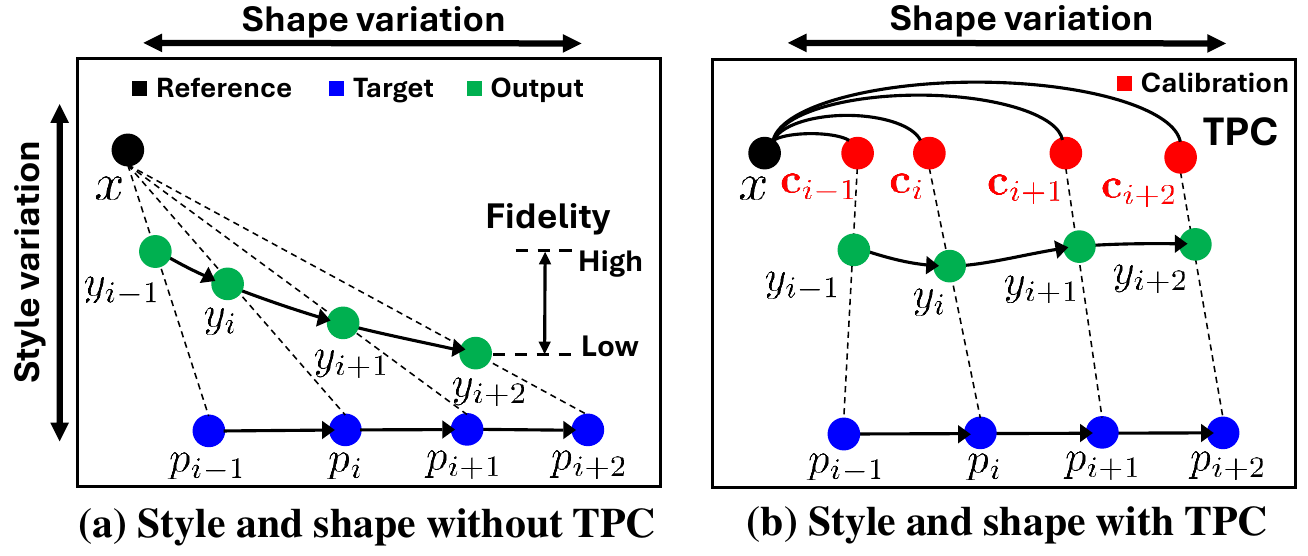}
    \caption{Conceptual illustration of the effectiveness of TPC in terms of style and shape variation.}
    \label{fig:effect_of_pc}
    \vskip -0.1in
\end{wrapfigure}
%
%
To this end, we propose a diffusion guidance referred to as Test-time Procrustes Calibration (TPC).
%
%
As shown in Figure \ref{fig:pc} (a), the existing diffusion-based human image animation system takes inputs of reference image $x$ and $i$-th target pose frame $p_{i}$, where it generates $i$-th output animation frame $y_{i}$. 
%
%
Here, the $x$ is encoded into the latent feature $z$, which serves as a conditioning input for the denoising diffusion model (\ie UNet).
As depicted in Figure \ref{fig:pc} (b), our proposed TPC incorporates an auxiliary branch into this diffusion conditioning process.
This branch is defined as a calibration branch that guides denoising UNet to properly capture visual correspondence between target and reference.
%
To be specific, the TPC provides a calibrated reference image latent $\mathbf{c}_{i}$ that aligns with the human shape in the target pose $p_{i}$ based on statistical shape analysis referred to as Procrustes analysis \cite{gower1975generalized}. 
Figure \ref{fig:effect_of_pc} offers a qualitative understanding of the influence of this $\mathbf{c}_i$.
%
Conceptually, in Figure \ref{fig:effect_of_pc} (a), when the humans in reference image $x$ and the target pose sequence $p = \{p_{i-1}, p_{i}, p_{i+1}, p_{i+2}\}$ are projected onto the ideal 2D shape-style space, they show distinct locations in terms of style axis, where the system aims to generate image animation frame $y=\{y_{i-1}, y_{i}, y_{i+1}, y_{i+2}\}$ with the style of $x$ and the shape of $p$. 
%
%
%
%
However, in cases (\eg $x$ and $p_{i+2}$) where significant gaps exist along the shape axis (\ie compositional misalignment), the current diffusion model struggles to preserve the original style. 
This leads to low-fidelity outputs (\eg $y_{i+2}$) and results in temporal inconsistencies due to unstable fidelity across frames.
Thus, as illustrated in Figure \ref{fig:effect_of_pc} (b), we bridge this gap in shapes between reference and target by providing correspondence guidance latent $\mathbf{c}$ by our designed TPC.
With this guidance condition, diffusion-based animation systems achieve robustness to fidelity variations and maintain temporal consistency among frames.
The TPC is simple and works in a model-agnostic manner without additional training and validates its effectiveness on human image animation benchmarks (\ie TikTok\cite{jafarian2021learning}, TED-talks \cite{siarohin2021motion}) and even in unseen domain data of real environment scenarios.

\section{Related Work}
\label{related_work}

\subsection{Diffusion-based Human Image Animation}
Human image animation aims to provide a video about an animated version of an input human image.
%
%
As a human-centric application of image-to-video technology, the human image animation systems \cite{siarohin2021motion,siarohin2019first,zhao2022thin} have previously been developed based on generative adversarial networks.
%
%
The recent emergence of denoising diffusion models \cite{song2020score,song2020denoising,ho2020denoising} has presented a new paradigm for generative models, where image animation has also faced fundamental innovations.
The diffusion-based framework generates an image animation from noise using pre-trained denoising capabilities based on input human image and target motion video, where the motion video can be extracted from various pose estimation models \cite{cao2017realtime,guler2018densepose}.
%
Early work of diffusion-based image animation was made in DreamPose \cite{karras2023dreampose}, which leveraged the pre-trained text-to-image diffusion model (\eg Stable Diffusion \cite{rombach2022high}) by conditioning on human image embeddings instead of text, rendering videos of humans in various outfits performing simple walking motions.
AnimateAnyone \cite{hu2023animate} introduces a UNet-style reference image encoder that enhances the layered conditioning of the reference image following the encoding-decoding process of UNet.
Furthermore, ControlNet \cite{zhang2023adding} has been a popular choice with the diffusion model by offering more controlled guidance about target motion.
%
To enhance background fidelity, DisCo \cite{wang2023disco} segments the background of the reference image and integrates it into ControlNet, along with the target motion.
%
For the temporal consistency of output video, MagicAnimate \cite{xu2023magicanimate} introduces temporal attention by inflating the original 2D UNet to 3D temporal UNet.
However, current systems still suffer from quality irregularity issues when the human shapes are not aligned between the reference image and the target motion.
Such misalignments frequently occur in real-world scenarios, prompting our proposed Procrustes calibration to address this challenge.
\section{Preliminary}
\subsection{Procrustes Analysis}
Procrustes\footnote{The name ``Procrustes'' comes from Greek mythology, where Procrustes was a bandit who would stretch or cut people to fit his bed, reflecting the idea of adjusting data to fit a common form.} analysis (PA) \cite{gower1966some,gower1975generalized} is a statistical shape analysis technique used to compare the shapes of objects \cite{hong2023joint,martinez2017simple}. 
The PA involves finding the best alignment between two shapes by scaling, translating, and rotating one shape to match the other as closely as possible.
To formulate the process of PA, we are given two sets of $n$ points as $X = \{x_{1},\cdots,x_{n}\} \in \mathbb{R}^{n \times d}$ and $Y = \{y_{1},\cdots,y_{n}\} \in \mathbb{R}^{n \times d}$, where $d$ is the dimension of the point.
We perform Procrustes transformation composed of scaling factor $s \in \mathbb{R}^{1}$, rotation matrix $r \in \mathbb{R}^{d \times d}$, and translation vector $t \in \mathbb{R}^{1 \times d}$ as given below:
\begin{equation}
\begin{aligned}
\hat{Y} = s\cdot X r + t,
\end{aligned}
\label{eq:pa}
\end{equation}
where the vector $t$ is added with broadcasting to all $n$ points.
The objective is to find the optimal transformation parameters $s, r, t$ to minimize the sum of squared differences between $\hat{Y}$ and the $Y$.
\begin{equation}
\begin{aligned}
\underset{s,r,t}{\mathrm{argmin}} ||\hat{Y} - Y||_{F},
\end{aligned}
\label{eq:pa_obj}
\end{equation}
where $||\cdot||_{F}$ is Frobenius norm.
The optimal value $r^{*}$ is typically obtained via singular value decomposition and the $s^{*}$ and $t^{*}$ are calculated after the optimal rotation $r^{*}$ is found.
Here, we apply this PA to align the human shapes of a reference image and target motion frame.
\section{Method}
\begin{figure*}[t!]
\centering
    \vskip -0.15in
   \includegraphics[width=1.0\textwidth]{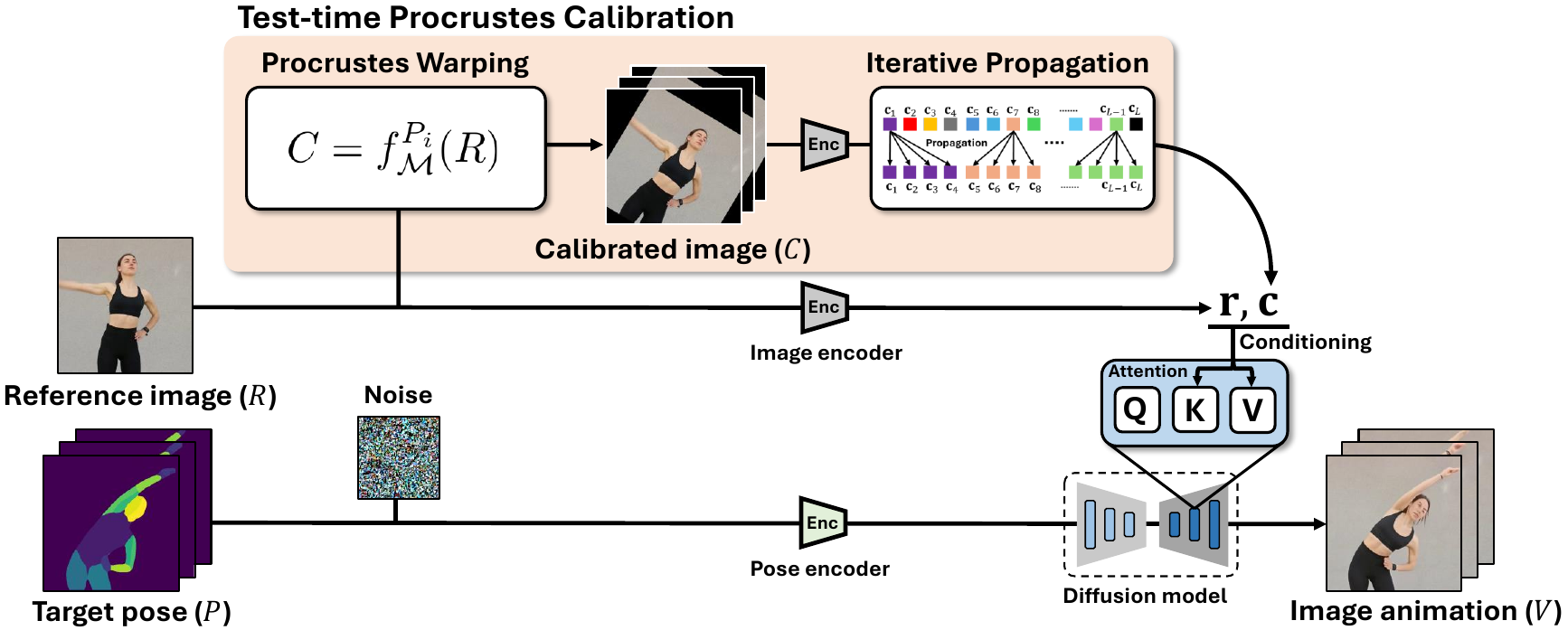}
   \caption{Illustration of Test-time Procrustes Calibration (TPC) on diffusion-based human image animation. TPC provides calibrated latent feature $\mathbf{c}$ to enhance shape correspondence between the reference image and target poses. Procrustes Warping aligns the reference image with the target pose shape, while Iterative Propagation improves temporal consistency among calibrated features.}
\label{fig:model}
\vskip -0.15in
\end{figure*}

%
%
%
%
%
Given a human reference image $R$ and a target pose sequence $P = [P_{1}, \cdots, P_{L}]$, a human image animation system generates image animation video $V = [V_{1}, \cdots, V_{L}]$ which follows the pose sequence by the human in the reference image, where $L$ is the number of frames.
%
%
Figure \ref{fig:model} shows the application of Test-time Procrustes Calibration (TPC) into the general diffusion-based human image animation system.
%
%
%
The TPC aims to improve the quality of resulting animation by consistently ensuring the compositional alignment of human shapes between the reference image and the target poses.
%
At each denoising step $t$, TPC takes an input reference image $R$ and target poses $P$, and produces calibrated image $C$ and its embedded latent $\mathbf{c}$ for conditioning in diffusion denoising.
The calibrated latent $\mathbf{c}$ guides the conditioning module (\ie cross-attention) in denoising UNet with precise correspondence about human shapes between the reference and the target poses.
%
%
%
%
To perform this, the TPC follows the sequential process of $R \rightarrow C \rightarrow \mathbf{c}$, and it comprises two main modules: (1) Procrustes Warping (Sec \ref{sec:4_1}) and (2) Iterative Propagation (Sec \ref{sec:4_2}).
%
%
%
Using the input reference image $R$ and target poses $P$, Procrustes warping produces a calibrated reference image $C$ optimized to align the human shape with each target pose.
%
%
%
After embedding this calibrated image $C$ into calibrated latent feature $\mathbf{c}$, Iterative Propagation iteratively refines the $\mathbf{c}$ during denoising step to enhance temporal consistency among the features by applying our designed feature propagation method. 
This calibrated latent $\mathbf{c}$ is finally given to diffusion model's conditioning module (\ie cross-attention) as a condition by concatenating with the original latent feature $\mathbf{r}$ of the reference image.
\subsection{Procrustes Warping}
\label{sec:4_1}
\begin{wrapfigure}{r}{0.5\textwidth}
    \centering 
    \vskip -0.2in
    \includegraphics[width=0.5\textwidth]{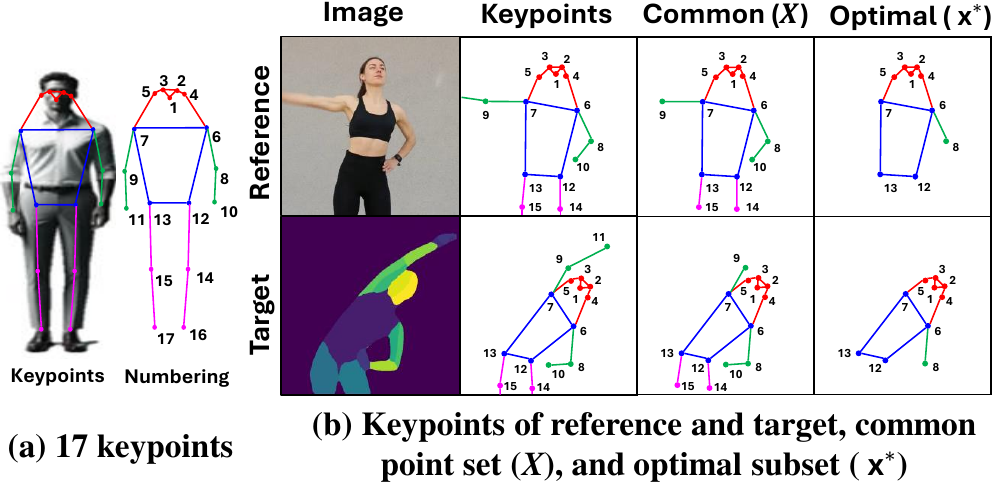}
    \vskip -0.05in
    \caption{Illustration of (a) pre-defined 17 keypoints and (b) keypoints in reference and target human, their common point set, and optimal set.}
    \label{fig:keypts}
    \vskip -0.1in
\end{wrapfigure}
%
Procrustes Warping (PW) aims to align human shapes between reference and target pose.
%
%
Thus, the PW takes a reference image $R$ as input and produces a calibrated reference image $C =[C_{1},\cdots,C_{L}]$ to the target human pose $P=[P_{1},\cdots,P_{L}]$.
%
%
To construct $C$, we first extract keypoint sets from both the reference and target humans. 
Then, we apply Procrustes analysis \footnote{Reason for choosing the Procrustes transform: Transformations are largely categorized into (1) shape-preserving (\eg linear, Procrustes) and (2) shape-distorting (\eg affine). To put the conclusion first, the Procrustes transform was the most effective. Shape-distorting methods were less effective than shape-preserving due to the loss of visual information by distortion. Detailed analysis is available in Table \ref{tab:absty} and Figure \ref{fig:cal}.} between the two sets, determining transformation parameters: scaling, rotation, and translation. 
%
%
%
%
%
%
%
%
%
Using these parameters, we transform the reference image to align it with the target.
%
%
%
To be specific, as shown in Figure \ref{fig:keypts} (a), we define 17 keypoints \footnote{We place more points on informative regions of human identity (face: 5, torso: 4, arms: 4, and legs: 4).} in 2-dimensional space for human body and face using keypoint extractor (\eg OpenPose \cite{cao2017realtime}). 
%
%
%
%
Figure \ref{fig:keypts} (b) illustrates that we obtain these keypoints from the reference image $R$ and $i$-th target pose frame $P_{i}$. 
From these, we filter out commonly visible $N$ points, defining $X =\{x_{1},\cdots,x_{N}\}$ as reference set and $Y =\{y_{1},\cdots,y_{N}\}$ as target set. 
Following Procrustes analysis (\ie Eq. (\ref{eq:pa},\ref{eq:pa_obj})), we obtain optimal transformation parameters $s^{*} \in \mathbb{R}^{1}, r^{*} \in \mathbb{R}^{2 \times 2}, t^{*} \in \mathbb{R}^{1 \times 2}$ and warp all pixels $[u,v] \in \mathbb{R}^{n \times 2}$ ($n$ is the number of pixels) in the reference image by mapping as below: 
\begin{equation}
\begin{aligned}
\mathcal{M}:[u,v] \rightarrow s^{*}[u,v] \cdot r^{*} + t^{*}.
\end{aligned}
\label{eq:mapping}
\end{equation}
Therefore, we obtain the $i$-th calibrated image $C_{i} = f_{\mathcal{M}}^{P_{i}}(R)$ using Procrustes warping $f_{\mathcal{M}}^{P_{i}}$, which aligns $R$ with human shape in $P_{i}$ using mapping $\mathcal{M}$.
%
%
However, considering all $N$ common visible points for warping is unreasonable since the reference and target shapes cannot perfectly overlap due to differing poses (\eg the arms in Figure \ref{fig:keypts} (b) cannot overlap).
%
%
%
%
%
To address this, we apply Procrustes warping to subset $\mathbf{x} \subset X$ of the common points and select the most effective subset $\mathbf{x}^{*}$ based on our defined alignment score function $h$ as below (subset $\mathbf{y}$ is also defined corresponding to the $\mathbf{x}$):
\begin{equation}
\begin{aligned}
\mathbf{x}^{*} = \underset{\mathbf{x}}{\textrm{argmax}} \hspace{0.1cm} h(P_{i}, C_{i}^\mathbf{x}),
\end{aligned}
\label{eq:sup}
\end{equation}
where $C^{\mathbf{x}}_{i}$ denotes calibrated image using keypoints of subset $\mathbf{x}$.
%
%
The $h$ is alignment score function that computes pixel-wise IoU\footnote{SAM \cite{kirillov2023segment} is used for segmenting each human shape to measure pixel-wise IoU.} of human shapes between pose $P_{i}$ and calibrated image $C_{i}^\mathbf{x}$.
Thus, the final $i$-th calibrated image is defined as $C_{i} = C_{i}^{\mathbf{x}^{*}}$ with optimal subset $\mathbf{x}^{*}$.
\subsection{Iterative Propagation}
\label{sec:4_2}
\begin{wrapfigure}{r}{0.38\textwidth}
    \centering
    \vskip -0.2in
    \includegraphics[width=0.38\textwidth]{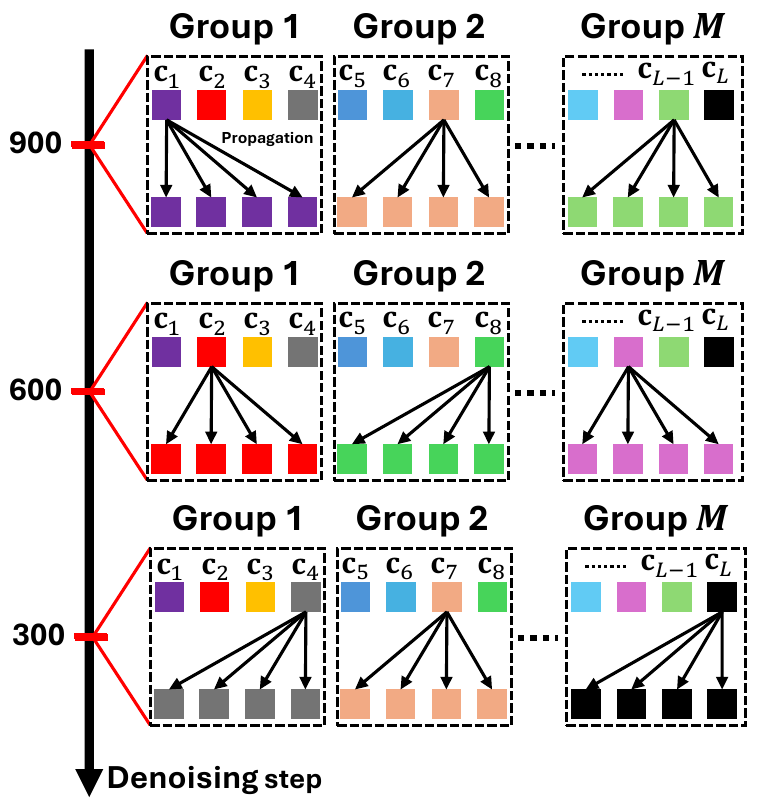}
    \caption{Illustration of iterative propagation on calibrated latent features. It shows $M$ groups on $L$ frame calibrated features and updates features in each group with randomly selected ones during the denoising process.}
    \label{fig:ip}
    \vskip -0.1in
\end{wrapfigure}
%
Conceptually, as shown in Figure \ref{fig:model}, our proposed TPC  aims to mitigate compositional misalignment by feeding calibrated image latent features $\mathbf{c} = [\mathbf{c}_{1},\cdots,\mathbf{c}_{L}]$ into the conditioning module (\ie cross-attention) of denoising diffusion along with the original reference latent feature $\mathbf{r}$.\footnote{The dimensions are $\mathbf{c} \in \mathbb{R}^{L \times m \times d}$ and $\mathbf{r} \in \mathbb{R}^{m \times d}$, by $d$-dimensional patch-wise image encoder $\textrm{Enc}(\cdot)$, where $m$ is the number of image patches and the $L$ is the number of frames.}
%
%
Thus, the $\mathbf{c}$ is designed to bridge the correspondence of the human shapes between the reference image and target pose.
%
%
However, the pose variation within the target pose frames affects the degree of calibration with the reference image, which reduces temporal consistency in the output.
To address this, we introduce Iterative Propagation (IP) to enhance consistency among calibrated latent features.
Figure \ref{fig:ip} illustrates the IP process during the diffusion denoising.
The IP forms $M$ groups of sequential features in calibrated features $\mathbf{c}$, randomly selects a feature within each group, and updates all features in the group with the selected one. 
%
%
This method enhances temporal consistency among calibrated latent features while maintaining compositional alignment with the target pose due to the continuity of target pose variation.
%
%
%
%
%
The random selection of features ensures all features have an equal chance of being chosen during the denoising process.
\subsection{Plug-and-Play Test-time Procrustes Calibration}
\label{pap}
We integrate calibrated latent feature $\mathbf{c}$ into human image animation systems by applying it into a conditioning module (\ie cross-attention) of video diffusion UNet. 
%
%
For $i$-th frame $m$ patch-wise image feature $\mathbf{p}_{i} \in \mathbb{R}^{m \times d}$, reference feature $\mathbf{r} \in \mathbb{R}^{m \times d}$, and calibrated latent feature $\mathbf{c}_{i} \in \mathbb{R}^{m \times d}$ the cross attention is formulated as $Q \leftarrow \textrm{Softmax}(Q K^{T}/d)V$, where it satisfies $Q = \mathbf{p}_{i}$ and $K = V = [\mathbf{r}, \mathbf{c}_{i}] \in \mathbb{R}^{2m \times d}$ is concatenated latent condition.
\section{Experiments}
\label{others}

\subsection{Experimental Settings}
\paragraph{Implementation Details.}
SAM \cite{kirillov2023segment} is used for screening out the background in calibrated images.
VQ-VAE \cite{van2017neural} is used for encoding images of the video.
The number of groups in iterative propagation is chosen as $M=30$ under ablation studies in Table \ref{tab:absty}.
The average number of video frames is about 120.
We use Stable Diffusion 1.5 \cite{rombach2022high} for all baselines on 4 NVIDIA A100 GPUs.
We follow the same pose encoders and image encoders of baseline models.
%
%
%
\paragraph{Data and Baselines.}
We validate human image animation on two popular benchmarks (\ie TikTok \cite{jafarian2021learning}, TED-talks \cite{siarohin2021motion}) about a test split.
Due to no validation splits, we provide valid sets matching the test set sizes for the ablation study.
%
We further collected 114 samples \footnote{Supplementary provides all the video links of these samples.} from TikTok and TED-talks as another test split. 
These samples contain compositional misalignment about rotation and scaling between human shapes in reference images and motion videos. 
%
%
The criteria for misalignment include relative angle and scaling differences, with samples having a relative angle $\theta$ > $\pi/6$ or relative scale IoU < 0.7.
As shown in Figure \ref{fig:observation_evidence} (a), the relative angle measures an angle between the straight lines from the pelvis to the nose in humans of reference and target poses using keypoints estimator \cite{cao2017realtime}.
The relative scaling measures the IoU (Intersection of Union) between bounding boxes of humans in the reference and the target.
Procrustes Calibration is validated on recent diffusion-based human image animation models including MagicAnimate \cite{xu2023magicanimate}, DisCo \cite{wang2023disco}, AnimateAnyone\footnote{As the code is not available, we use the work at: https://github.com/MooreThreads/Moore-AnimateAnyone} \cite{hu2023animate}, DreamPose \cite{karras2023dreampose} on their public codes and papers.
\subsection{Evaluation Metrics}
We evaluate videos in terms of single-frame quality and video quality.
For the single-frame, we measure Peak Signal-to-Noise Ratio (PSNR) \cite{hore2010image}, Structural Similarity Index Measure (SSIM) \cite{wang2004image}, Learned Perceptual Image Patch Similarity (LPIPS) \cite{zhang2018unreasonable}, FID (Fréchet Image Distance) \cite{heusel2017gans}, and L1 error between output and ground-truth images.
For the video quality, we measure Fréchet Video Distance (FVD) \cite{unterthiner2018towards} and FID-VID \cite{balaji2019conditional}.  
All automatic metrics are averaged over 10 runs with different seeds. We also analyze human preferences for results from the baselines with and without our TPC.

\begin{table}[t]
\scriptsize
  \caption{Quantitative evaluations of Test-time Procrustes Calibration (TPC) with recent diffusion-based human image animation models. A-Anyone: AnimateAnyone, M-Animate: MagicAnimate. It is reported in a format of (original test set / compositional misalignment test set).}
  \centering
  \begin{tabular}{l ccccc c cc c}
    \toprule
            \multirow{2}{*}{Method} &\multicolumn{5}{c}{Image} & &\multicolumn{2}{c}{Video} &\multirow{2}{*}{Human} \\ \cline{2-6} \cline{8-9}
		  & $L1 {\downarrow}_{\times \textrm{E-04}}$ & PSNR ↑ & SSIM ↑  & LPIPS ↓ & FID ↓ && FID-VID ↓ & FVD ↓ \\ 
    \midrule

\rowcolor{gray!30}\multicolumn{10}{c}{\textbf{TikTok \cite{jafarian2021learning}}} \\
\midrule
    DreamPose       & 7.22/9.64 & 27.31/25.17 & 0.532/0.481 & 0.449/0.529 & 55.4/87.2 && 61.1/93.1 & 568/738 &0.04 \\
    DreamPose  + TPC & 5.15/5.31 & 28.47/28.01 & 0.620/0.613 & 0.406/0.412 & 48.4/49.3 && 54.7/56.3 & 426/441 &0.96 \\ \hline
    DisCo        & 4.09/5.23 & 28.43/24.97 & 0.641/0.512 & 0.312/0.492 & 37.1/71.4 && 58.3/82.1 & 339/522 &0.28 \\
    DisCo + TPC   & 3.49/3.82 & 28.92/28.87 & 0.689/0.673 & 0.283/0.287 & 34.3/36.2 && 51.2/52.4 & 281/297 &0.72 \\ \hline
    A-Anyone           & 3.77/4.82 & 29.06/26.52 & 0.670/0.584 & 0.289/0.392 & 32.9/65.2 && 54.2/59.1 & 296/442 &0.24 \\
    A-Anyone + TPC      & 3.32/3.53 & 29.27/29.01 & 0.705/0.688 & 0.264/0.273 & 31.3/32.6 && 48.7/49.3 & 254/269 &0.76 \\ \hline
    M-Animate       & 3.17/4.36 & 29.11/27.82 & 0.717/0.641 & 0.241/0.321 & 31.8/49.2 && 22.4/52.3 & 182/362 &0.34 \\
    M-Animate + TPC  & \textbf{2.98}/3.19 & \textbf{29.43}/29.21 & \textbf{0.753}/0.731 & \textbf{0.232}/0.249 & \textbf{29.2}/30.4 && \textbf{21.0}/21.9 & \textbf{158}/164 &0.66 \\
\midrule

\rowcolor{gray!30}\multicolumn{10}{c}{\textbf{TED-talks \cite{siarohin2021motion}}} \\
\midrule
    DreamPose       & 6.65/7.41 & 27.63/26.11 & 0.559/0.482 & 0.421/0.521 & 63.4/86.2 && 43.6/64.2 & 411/532 &0.24 \\
    DreamPose + TPC  & 6.17/6.31 & 28.11/28.03 & 0.593/0.589 & 0.393/0.404 & 53.2/55.2 && 38.2/38.9 & 369/372 &0.76 \\ \hline
    DisCo        & 3.52/6.91 & 28.51/26.71 & 0.661/0.511 & 0.309/0.451 & 34.5/71.1 && 28.2/52.6 & 332/471 &0.24 \\
    DisCo + TPC   & 3.18/3.23 & 28.93/28.87 & 0.704/0.692 & 0.283/0.294 & 31.8/32.6 && 25.1/25.8 & 298/304 &0.76 \\ \hline
    A-Anyone           & 3.12/6.61 & 28.93/27.07 & 0.712/0.613 & 0.267/0.361 & 29.3/54.2 && 20.3/39.5 & 192/372 &0.24 \\
    A-Anyone + TPC      & 2.81/2.91 & 29.31/29.25 & 0.753/0.742 & 0.254/0.269 & 27.8/28.6 && \textbf{18.8}/19.5 & 173/178 &0.76 \\ \hline
    M-Animate       & 2.92/4.31 & 29.17/27.47 & 0.734/0.661 & 0.239/0.312 & 25.7/47.1 && 20.2/39.1 & 136/331 &0.24 \\
    M-Animate + TPC  & \textbf{2.77}/2.87 & \textbf{29.52}/29.41 & \textbf{0.782}/0.764 & \textbf{0.233}/0.249 & \textbf{24.3}/25.6 && 19.2/20.4 & \textbf{128}/137 &0.76 \\
    \bottomrule
  \end{tabular}

  \label{tab:1}
  \vskip -0.2in
\end{table}
\subsection{Experimental Results}
\paragraph{Quantitative Comparisons.}
Table \ref{tab:1} presents evaluations of image animation on two benchmark datasets (\ie TikTok, TED-talks) using recent diffusion image animation baselines (MagicAnimate, DisCo, AnimateAnyone, DreamPose) with TPC across three assessments (\ie image, video, human). Evaluations are conducted on two test splits: the original set and the compositional misalignment set. Initially, all baselines use a single reference image latent for conditioning all poses. After integrating TPC, they use calibrated latents corresponding to each pose.
%
Consistent improvements in image and video quality are observed in both test sets. All baselines struggle with the compositional misalignment set, but when integrated with TPC, they achieve quality close to the original test set. This demonstrates that morphological similarity affects the current diffusion model's ability to map human shapes from reference to target.
\begin{figure*}[t!]
\centering
   \includegraphics[width=1.0\textwidth]{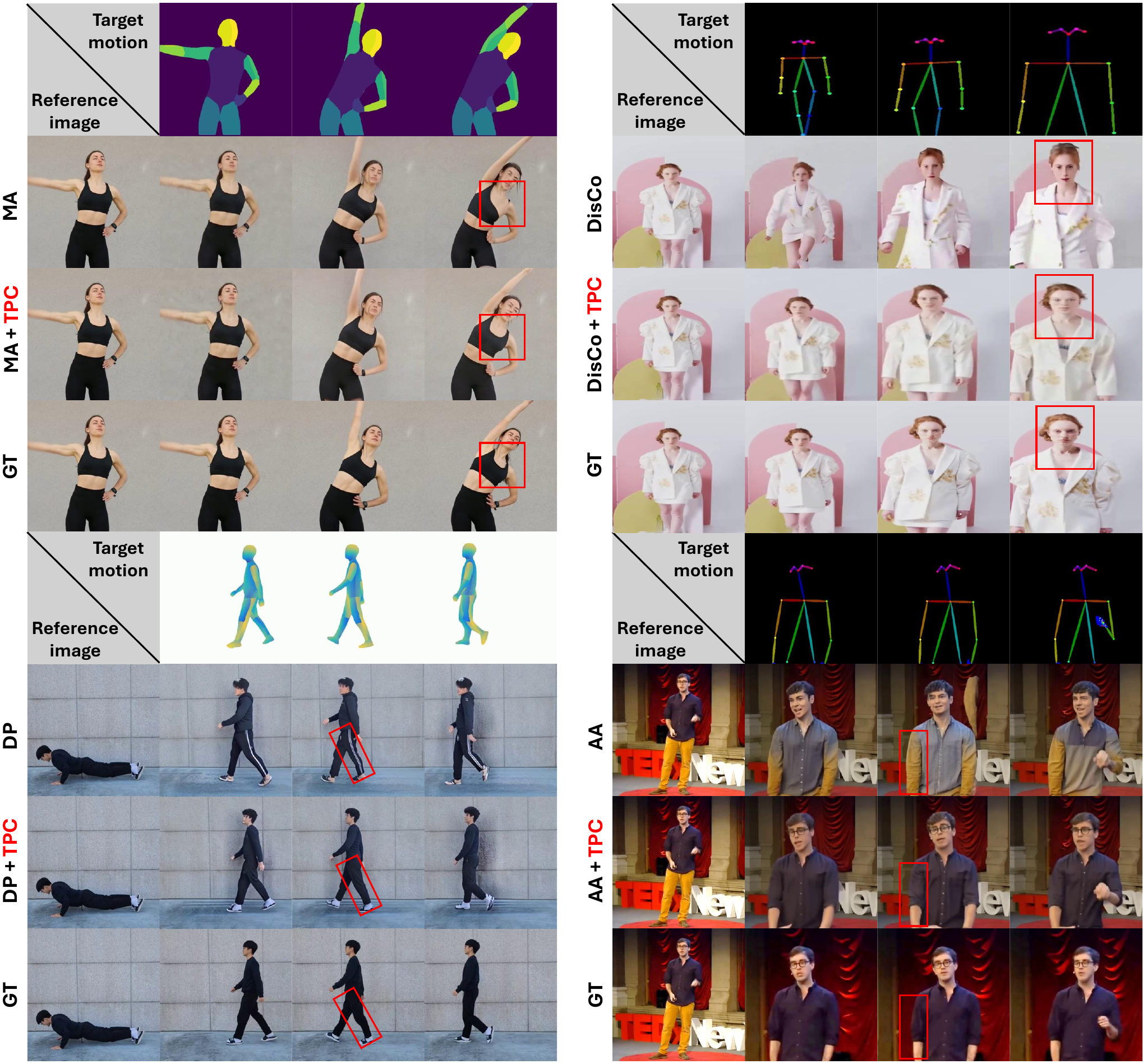}
   \caption{Qualitative results about applying TPC on diffusion-based human image animation systems (\ie MA: MagicAnimate, DisCo, AA: AnimateAnyone, DP: DreamPose) on cases about compositional misalignment of human shape between a reference image and target motion: Temporal misalignment by motion affecting factor rotation (top left) and scale (top right). Consistent misalignment affecting rotation (bottom left) and scale (bottom right). Calibrated images and predictions of compositional aligned samples are available in Appendix. Please see also video in the supplementary.}
\label{fig:qual}
\vskip -0.27in
\end{figure*}
\paragraph{Qualitative Comparisons.}
To validate our proposed TPC, we applied it to four recent baselines and compared the original results. We used the same types of human pose inputs (e.g., OpenPose, DensePose) for each baseline to prepare target motion videos. Figure \ref{fig:qual} shows predictions on four different samples exhibiting compositional misalignment.
%
%
The top left results show predictions on temporal misalignment between the reference and target due to the target's bending motion. MagicAnimate's fidelity diminishes with increased bending (red box in the last frame), whereas the model with TPC maintains high fidelity. The top right results display predictions on temporal misalignment due to a walking motion towards the front, with TPC similarly enhancing DisCo's performance.
%
%
The bottom left results show consistent misalignment across all frames. DreamPose struggles with low fidelity, causing unwanted stripes on the pants (red box), which are clearly removed with TPC.
The results in the bottom right exhibit consistent misalignment due to a scale difference. 
In AnimateAnyone, the reference human's yellow pants were incorrectly mapped onto the target human's arms, making them appear yellow.
However, the TPC completely mitigates this incorrect mapping.
This occurs because the calibrated image filters out the pants and enhances the correspondence of each body part. (Please, refer to the calibrated images also in the Appendix.)
\begin{wrapfigure}{r}{0.45\textwidth}
    \centering
    \includegraphics[width=0.45\textwidth]{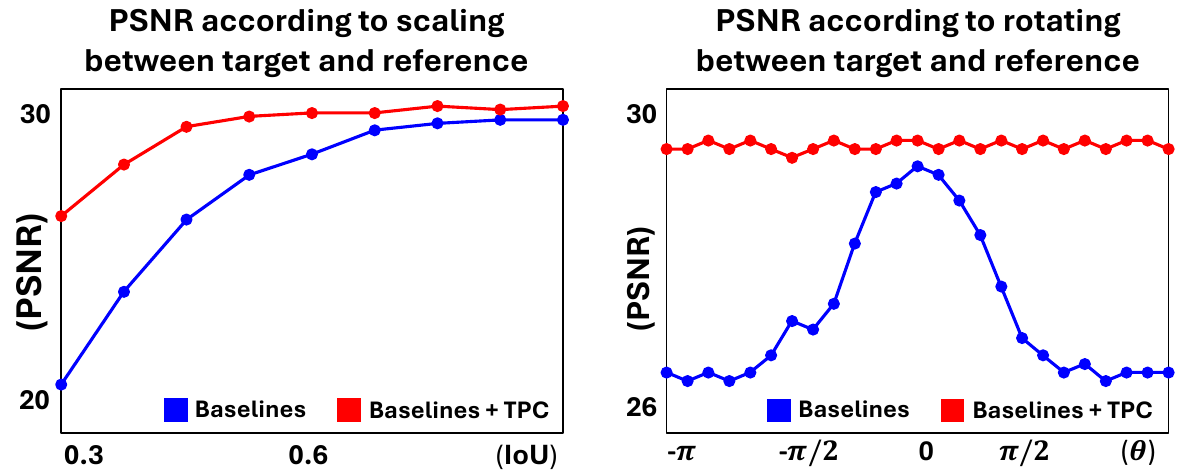}
    \caption{Robustness analysis of variations in compositional misalignment. Measurements of scale and rotation are in Figure \ref{fig:observation_evidence} (b).}
    \label{fig:q3}
    \vskip -0.1in
\end{wrapfigure}
Figure \ref{fig:qualitative2} shows the results on reference images from an unseen domain, generated by the T2I model \cite{ramesh2021zero}, applied in a compositional misalignment scenario. The left displays a temporal misalignment sample due to bending motions, with improved fidelity in both baselines, especially in DisCo.
The right side displays a consistent misalignment sample, where the baselines show low precision in transferring identity when the reference is flipped upside down. However, with the integration of TPC, their performance is significantly improved.
%
\vskip -0.3in
\paragraph{Robustness analysis} To assess the robustness of baselines with TPC, we measured fidelity (PSNR) by varying the scale and rotation of the human reference, as shown in Figure \ref{fig:q3}.
%
%
In the case of scaling, the baselines show a drop in performance when the size difference between the reference and target human shapes falls below an IoU of 0.6. However, TPC maintains optimal performance even down to an IoU of 0.4.
%
%
Notably, TPC enhances baseline robustness to all rotational variations by ensuring the calibrated image aligns well with shape differences caused by rotation.
\begin{figure*}[t!]
\centering
   \includegraphics[width=1.0\textwidth]{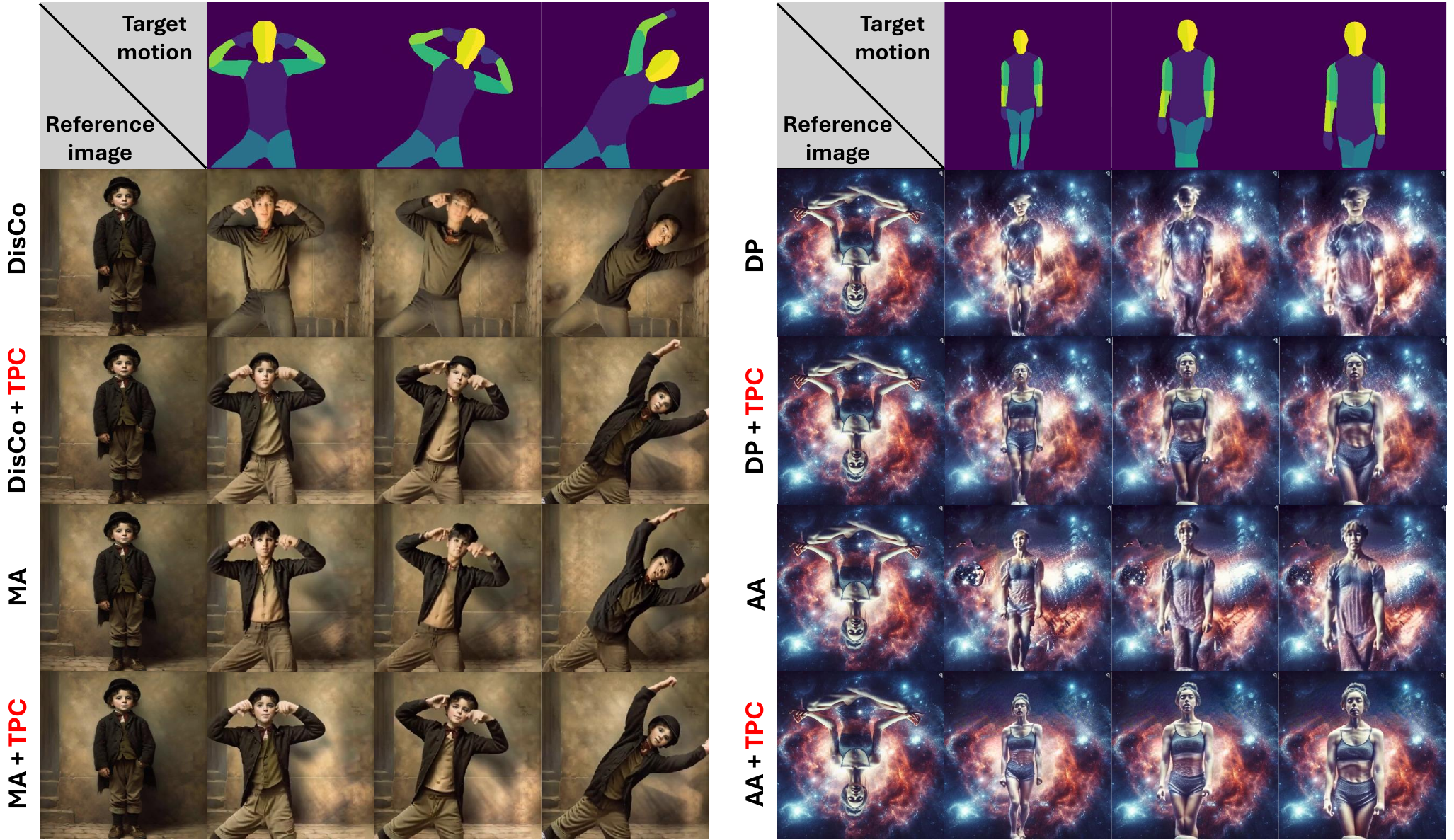}
   \caption{Qualitative results about unseen domain reference image. TPC enhances human image animation models' robustness on samples about compositional misalignment in the unseen domain.}
\label{fig:qualitative2}
\end{figure*}

\begin{figure}
\scriptsize
  \begin{minipage}[t]{.45\linewidth}
    \centering
    \captionof{table}{Ablation studies on transformation methods for reference image calibration and iterative propagation (IP) on TED-talks and TikTok. (validation splits, average score compositional alignment/misalignment).}
    \begin{tabular}{lcccc}
    \toprule
    \multirow{2}{*}{Method} &\multicolumn{2}{c}{Foreground} & \multicolumn{2}{c}{Background} \\ \cline{2-5}
    & SSIM↑  & FVD↓ & SSIM↑ & FVD↓ \\
    \midrule
    Linear & 0.702 & 191 & 0.751 & 170   \\
    Affine & 0.704 & 193 & 0.754 & 171  \\
    Procrustes & \textbf{0.734} & \textbf{162} & \textbf{0.782} & \textbf{142} \\ \midrule
    w/o IP & 0.709 & 184 & 0.728 & 162 \\
    w/ IP ($M$=20) & 0.731 & 164 & 0.776 & 145  \\
    w/ IP ($M$=30) & \textbf{0.734} & \textbf{162} & \textbf{0.782} & \textbf{142} \\
    w/ IP ($M$=40) & 0.728 & 165 & 0.777 & 145 \\
    \bottomrule
    \label{tab:absty}
    \end{tabular}
  \end{minipage}\hfill 
  \begin{minipage}[t]{.23\linewidth}
  \centering
  \captionof{table}{Inference time of baselines with TPC. DC: DisCo.}
  \label{tab:inf_time}
  \begin{tabular}{lc}\\\toprule  
Method & sec/frame \\\midrule
DP  & 18.3 \\
DP + TPC & 18.9 \\ \midrule
DC  & 5.8 \\
DC + TPC & 6.3 \\ \midrule
AA  & 2.3 \\
AA + TPC & 2.6 \\ \midrule
MA  & 1.4 \\
MA + TPC & 1.6 \\  \bottomrule
\end{tabular}
  \end{minipage}\hfill 
  \begin{minipage}[t]{.27\linewidth}
    \centering
    \vspace{0.5mm} 
    \includegraphics[width=0.99\linewidth]{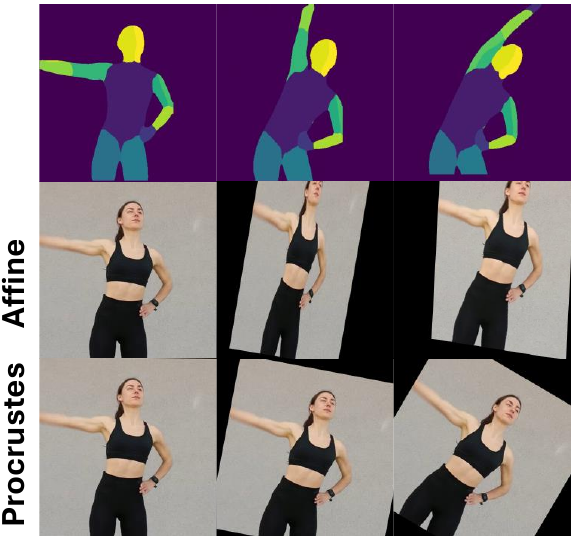}
    \captionof{figure}{Comparisons of calibrated images using an affine transform and Procrustes transform.}
    \label{fig:cal}
  \end{minipage}
  \vskip -0.2in
\end{figure}

\paragraph{Ablation Study.}
%
%
Table \ref{tab:absty} presents ablation studies on transformation methods and iterative propagation. Transformations are categorized as shape-preserving (\eg Linear, Procrustes) and shape-distorting (\eg Affine). Linear transformation performs scaling and rotation without translation, based on the object's bounding box coordinates. The Procrustes method is the most effective, transforming by selecting the optimal keypoint. The affine transform is less effective due to shearing, which can cause distortion and information loss by moving the subject out of the frame, as shown in Figure \ref{fig:cal}.
%
The second section presents an ablation study on iterative propagation, showing significant enhancements in foreground and background, with $M$=30 being the most effective.
Table \ref{tab:inf_time} provides inference time on baselines with TPC.
Our method requires only minimal additional time because it processes all frames batch-wise, generating the calibrated image in a single iteration.
%
%
\subsection{Broader Impacts and Ethic Statements}
Visual generative models present a range of ethical dilemmas, including the creation of unauthorized counterfeit content, potential privacy breaches, and challenges related to fairness.
Due to our reliance on the architecture of these models, our work inherently adopts these ethical vulnerabilities. 
Addressing these concerns is imperative and requires the establishment of comprehensive regulations and technical countermeasures.
%
%
We are exploring advanced measures such as learning-based digital forensics and digital watermarking to ethically navigate the complexities of visual generative models.
\subsection{Limitation and Future work}
We provide an overview of the various limitations and potential development directions identified during this study.
%
Human image animation systems transfer reference images to target poses. However, some frames in a target pose video may lack proper specification, leading to flicker or low fidelity in model predictions. Thus, integrating high-quality pose estimation is crucial.
%
Another limitation is that significant differences in body shape between the reference and target poses result in awkward transfers, such as transferring a skinny person to a fat target pose. 
To achieve natural results, a system or module that aligns body shapes is required. 
For the future extension of human image animation system, it is essential to ensure robust operations under multiple individuals.
The system should ensure consistent identity transfer across multiple individuals and accurately adapt to those appearing at specific times.
To achieve this, video technologies about recognition \cite{ravi2024sam,zhu2017flow} and perception \cite{yoon2022selective,yoon2023scanet} can be further incorporated.
Currently, conditioning relies solely on images, but it is anticipated that various other modalities, such as text and audio, could also be incorporated.
Especially, the integration of the text modality with emerging large language models \cite{brown2020language,touvron2023llama} is expected to drive innovative industrial advancements, with the introduction of video/image-based conversational systems \cite{yoon2024bi,yoon2022information,yoon2023hear} serving as a compelling utility.
Lastly, integrating technologies focused on speed \cite{koo2024wavelet,bolya2022token}, resource efficiency \cite{yang2023diffusion}, and test-time calibration \cite{yoon2023esd} is expected to improve human image animation systems' applicability into real-world environments.
%
%
\section{Conclusion}
%
%
Current diffusion-based human image animation systems are facing challenges on samples of compositional misalignment of human shapes between a reference image and target pose frames.
%
To this end, this paper presents Test-time Procrustes Calibration (TPC) which improves the robustness of image animation models on the compositional misalignment samples in a model-agnostic manner.
Extensive experiments demonstrate the effectiveness of TPC.
\section*{Acknowledgements}
This work was supported by Institute for Information \& communications Technology Planning \& Evaluation (IITP) grant funded by the Korea government(MSIT) (No.RS-2021-II211381, Development of Causal AI through Video Understanding and Reinforcement Learning, and Its Applications to Real Environments) and partly supported by Institute of Information \& communications Technology Planning \& Evaluation (IITP) grant funded by the Korea government(MSIT) (No.RS-2022-II220184, 2022-0-00184, Development and Study of AI Technologies to Inexpensively Conform to Evolving Policy on Ethics)
%
{
\small
\bibliographystyle{plain}
\bibliography{bibliography}
}

\clearpage
\appendix

\section{Appendix}
%
%
\subsection{Algorithm for Iterative Propagtion}
\begin{algorithm}[ht]
    \caption{Iterative Propagation}
    \label{alg:ip}
    \begin{algorithmic}[1]
        \State \textbf{Input}: Calibrated $L$ frame features $\mathbf{c} = [\mathbf{c}_{1}, \cdots, \mathbf{c}_{L}]$, Number of frame groups $M$
        \State \textbf{Output}: Denoising $t$-step propagated features $\mathbf{c}^{t} = [\mathbf{c}_{1}^{t}, \cdots, \mathbf{c}_{L}^{t}]$
        \State Initialize the number of frames in each group: $m = \left\lfloor \frac{L}{M} \right\rfloor$
        \State \textbf{for} $t = T$ \textbf{to} $1$ \textbf{do} \Comment $T$ is total denoising step.
        \State \quad \textbf{for} $i = 1$ \textbf{to} $M$ \textbf{do}
        \State \quad \quad $n \leftarrow (i-1) \times m$
        \State \quad \quad Sample $r \sim \mathcal{U}\{1, m\}$
        \Comment $\mathcal{U}\{1, m\}$ is a discrete uniform distribution between 1 and $m$.
        \State \quad \quad $[\mathbf{c}_{n+1}^{t}, \mathbf{c}_{n+2}^{t}, \cdots, \mathbf{c}_{n+m}^{t}] \leftarrow [\mathbf{c}_{n+r}, \mathbf{c}_{n+r}, \cdots, \mathbf{c}_{n+r}]$ \Comment Update list.
        \State \quad \textbf{end}
        \State \quad $\mathbf{c}^{t} \leftarrow [\mathbf{c}_{1}^{t}, \cdots, \mathbf{c}_{m}^{t}] \cup [\mathbf{c}_{m+1}^{t}, \cdots, \mathbf{c}_{2m}^{t}] \cup \cdots \cup [\mathbf{c}_{m(M-1)+1}^{t}, \cdots, \mathbf{c}_{L}^{t}]$
        \Comment Concatenation
        \State \textbf{end}
    \end{algorithmic}
\end{algorithm}
\subsection{More qualitative results}
\textbf{Explanation}

\begin{figure*}[h!]
\centering
   \includegraphics[width=1.0\textwidth]{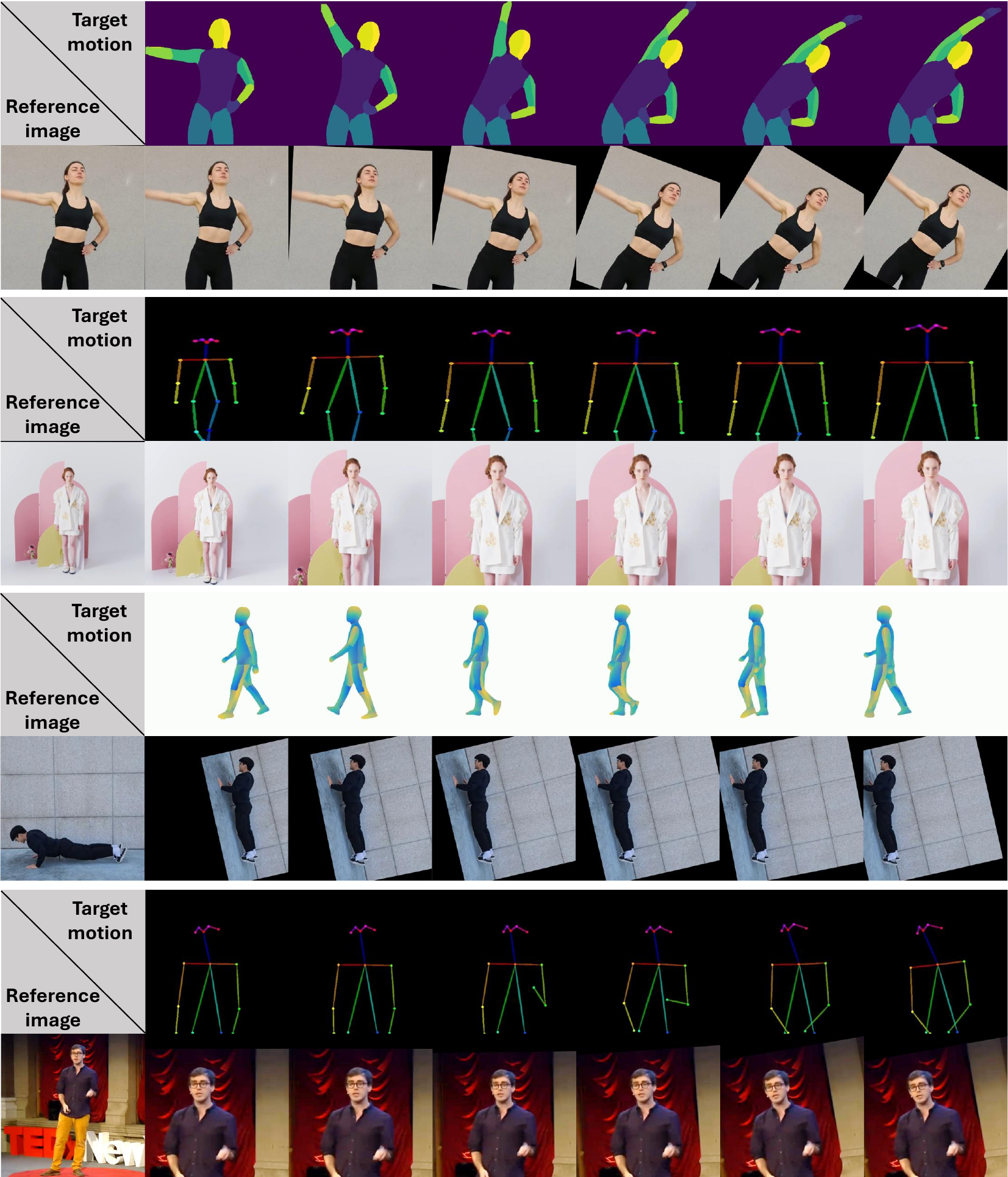}
   \caption{Qualitative results about calibrated images on the samples in Figure \ref{fig:qual}.}
\label{fig:page0}
\end{figure*}

Figure \ref{fig:page0} shows calibrated reference images corresponding to the samples in Figure 8.

\begin{figure*}[h!]
\centering
   \includegraphics[width=1.0\textwidth]{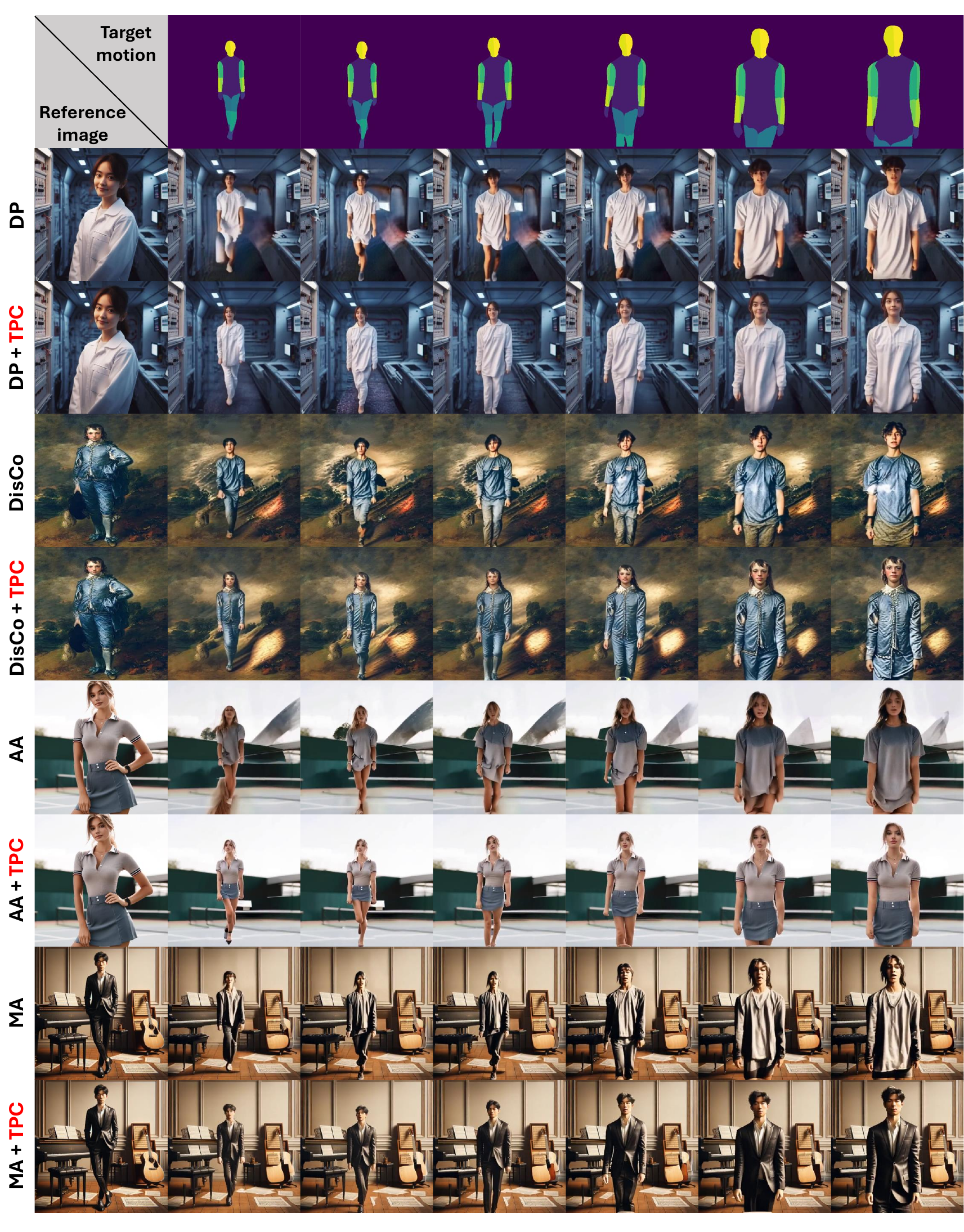}
   \caption{Qualitative results about applying TPC on recent diffusion-based human image animation systems (\ie DreamPose (DP),  MagicAnimate (MA), DisCo, AnimateAnyone(AA)) on identical motions with different reference images. The images are obtained from T2I image generation model.}
\label{fig:page1}
\end{figure*}

Figure \ref{fig:page1} shows results on identical motion with different reference images.

\begin{figure*}[h!]
\centering
   \includegraphics[width=1.0\textwidth]{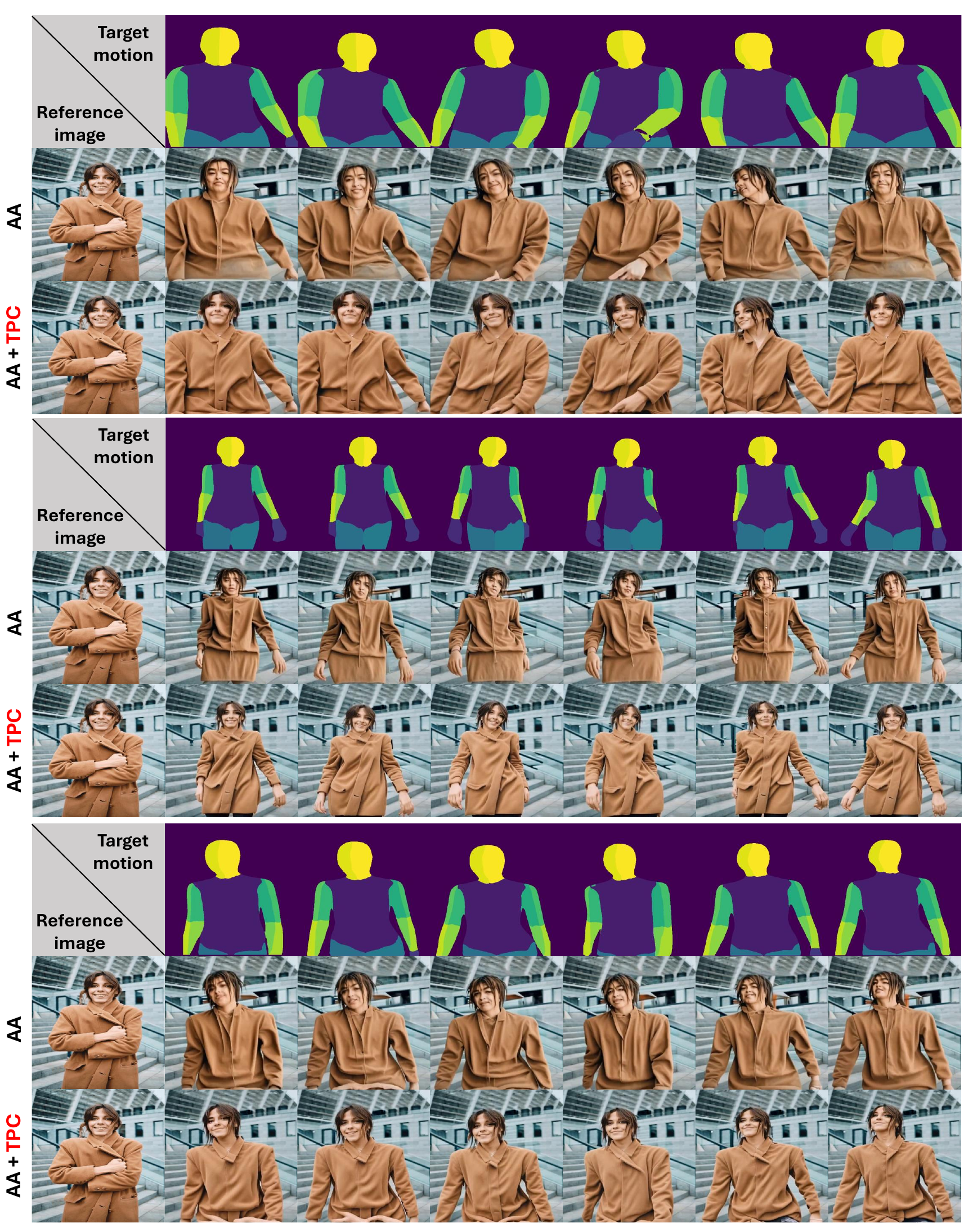}
   \caption{Qualitative results about applying TPC on recent diffusion-based AnimateAnyone (AA) on identical reference image with different motions. The motion videos are obtained from TikTok dance video dataset. (test-split)}
\label{fig:page2}
\end{figure*}

\begin{figure*}[h!]
\centering
   \includegraphics[width=1.0\textwidth]{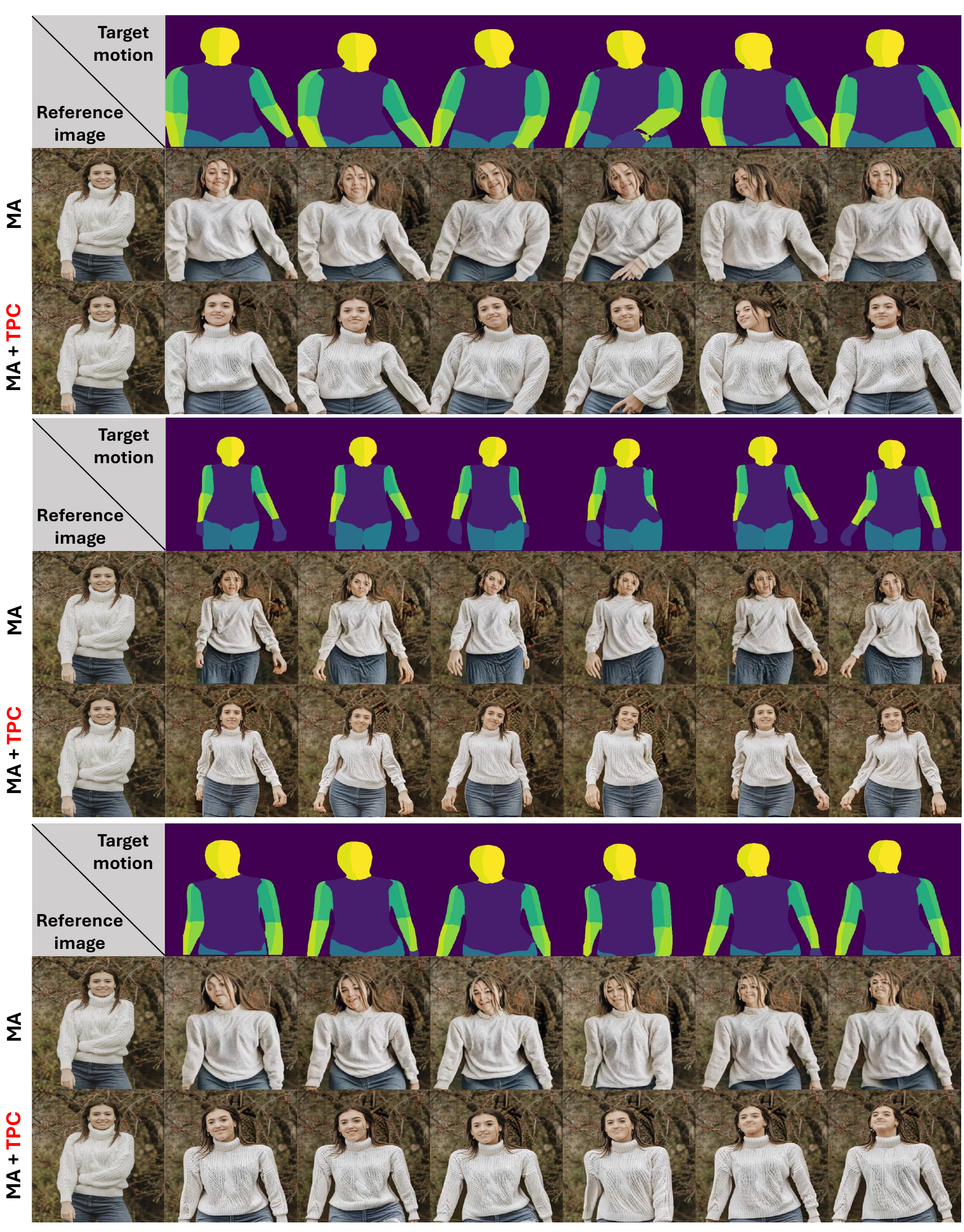}
   \caption{Qualitative results about applying TPC on recent diffusion-based MagicAnimate (MA) on identical reference image with different motions. The motion videos are obtained from TikTok dance video dataset. (test-split)}
\label{fig:page3}
\end{figure*}

Figure \ref{fig:page2} and \ref{fig:page3} show results on identical references with different motion videos in the TikTok dataset.

\begin{figure*}[h!]
\centering
   \includegraphics[width=1.0\textwidth]{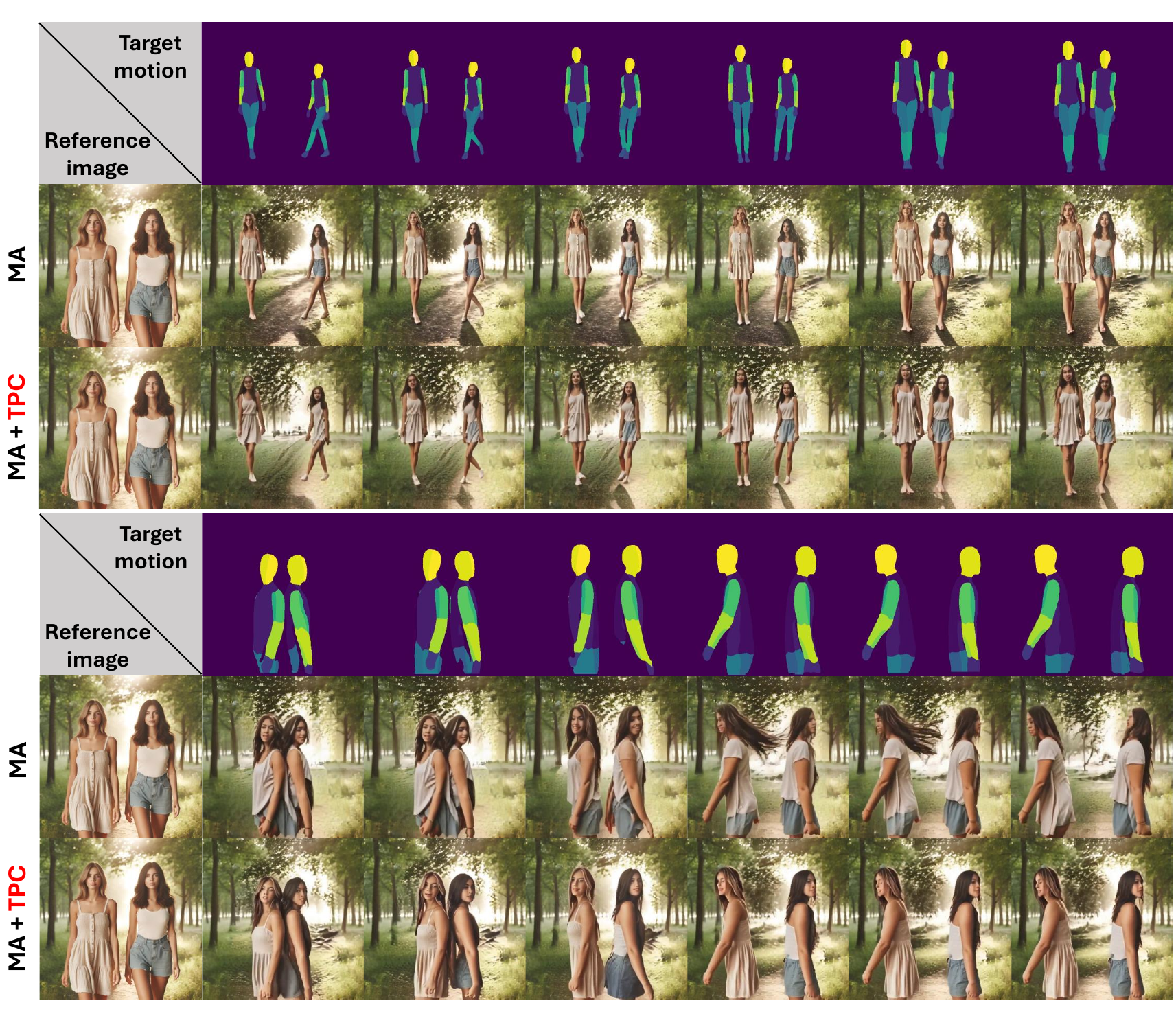}
   \caption{Qualitative results about applying TPC on recent diffusion-based MagicAnimate (MA) on multiple humans of reference and motion.}
\label{fig:page4}
\end{figure*}

Figure \ref{fig:page4} shows results on multiple humans in reference and target motion.

As in Figure 1, here we visualize the target pose in a format of DensePose for the visibility of target motions. The video samples in Figure 8 are publicly available from: 

- https://www.pexels.com/video/a-woman-stretching-5510095/

- https://www.pexels.com/video/modeling-wedding-dresses-7305163/

- https://www.youtube.com/watch?v=8S0FDjFBj8o

- https://www.tiktok.com/@dbstjswo505/video/7371773469899476231
\clearpage
\clearpage
\clearpage
\clearpage
\clearpage
\clearpage

\newpage
\section*{NeurIPS Paper Checklist}


\begin{enumerate}

\item {\bf Claims}
    \item[] Question: Do the main claims made in the abstract and introduction accurately reflect the paper's contributions and scope?
    \item[] Answer: \answerYes{} 
    \item[] Justification: Yes, in the line 15-17 in the Abstract and last paragraph in the Introduction.

\item {\bf Limitations}
    \item[] Question: Does the paper discuss the limitations of the work performed by the authors?
    \item[] Answer: \answerYes{} 
    \item[] Justification: It is provided as a Limitation section.

\item {\bf Theory Assumptions and Proofs}
    \item[] Question: For each theoretical result, does the paper provide the full set of assumptions and a complete (and correct) proof?
    \item[] Answer: \answerNA{} 
    \item[] Justification: \answerNA{}

    \item {\bf Experimental Result Reproducibility}
    \item[] Question: Does the paper fully disclose all the information needed to reproduce the main experimental results of the paper to the extent that it affects the main claims and/or conclusions of the paper (regardless of whether the code and data are provided or not)?
    \item[] Answer: \answerYes{} 
    \item[] Justification: Method section contains the details of our proposed framework.

\item {\bf Open access to data and code}
    \item[] Question: Does the paper provide open access to the data and code, with sufficient instructions to faithfully reproduce the main experimental results, as described in supplemental material?
    \item[] Answer: \answerYes{} 
    \item[] Justification: We submit the source links for data used in our experiment and also a demo video.

\item {\bf Experimental Setting/Details}
    \item[] Question: Does the paper specify all the training and test details (e.g., data splits, hyperparameters, how they were chosen, type of optimizer, etc.) necessary to understand the results?
    \item[] Answer: \answerYes{} 
    \item[] Justification: It is explained in the Implementation Details section.

\item {\bf Experiment Statistical Significance}
    \item[] Question: Does the paper report error bars suitably and correctly defined or other appropriate information about the statistical significance of the experiments?
    \item[] Answer: \answerYes{} 
    \item[] Justification: It is explained in the evaluation metrics.

\item {\bf Experiments Compute Resources}
    \item[] Question: For each experiment, does the paper provide sufficient information on the computer resources (type of compute workers, memory, time of execution) needed to reproduce the experiments?
    \item[] Answer: \answerYes{} 
    \item[] Justification: We provide information about GPU resources in the Implementation Details section for implementing our framework.
    
\item {\bf Code Of Ethics}
    \item[] Question: Does the research conducted in the paper conform, in every respect, with the NeurIPS Code of Ethics \url{https://neurips.cc/public/EthicsGuidelines}?
    \item[] Answer: \answerYes{} 
    \item[] Justification: We read the NeurIPS Code of Ethics and follow all of it.

\item {\bf Broader Impacts}
    \item[] Question: Does the paper discuss both potential positive societal impacts and negative societal impacts of the work performed?
    \item[] Answer: \answerYes{} 
    \item[] Justification: We provide a broader impact including societal negative impacts.
    
\item {\bf Safeguards}
    \item[] Question: Does the paper describe safeguards that have been put in place for responsible release of data or models that have a high risk for misuse (e.g., pretrained language models, image generators, or scraped datasets)?
    \item[] Answer: \answerYes{} 
    \item[] Justification: The broader impact contains information about safeguards.

\item {\bf Licenses for existing assets}
    \item[] Question: Are the creators or original owners of assets (e.g., code, data, models), used in the paper, properly credited and are the license and terms of use explicitly mentioned and properly respected?
    \item[] Answer: \answerYes{} 
    \item[] Justification: We explicitly provide all the references we used including url links.

\item {\bf New Assets}
    \item[] Question: Are new assets introduced in the paper well documented and is the documentation provided alongside the assets?
    \item[] Answer: \answerNA{} 
    \item[] Justification: \answerNA{}

\item {\bf Crowdsourcing and Research with Human Subjects}
    \item[] Question: For crowdsourcing experiments and research with human subjects, does the paper include the full text of instructions given to participants and screenshots, if applicable, as well as details about compensation (if any)? 
    \item[] Answer: \answerNA{} 
    \item[] Justification: \answerNA{}

\item {\bf Institutional Review Board (IRB) Approvals or Equivalent for Research with Human Subjects}
    \item[] Question: Does the paper describe potential risks incurred by study participants, whether such risks were disclosed to the subjects, and whether Institutional Review Board (IRB) approvals (or an equivalent approval/review based on the requirements of your country or institution) were obtained?
    \item[] Answer: \answerNA{} 
    \item[] Justification: \answerNA{}

\end{enumerate}

\end{document}